\documentclass[10pt,twocolumn,letterpaper]{article}

\usepackage{iccv}
\usepackage{times}
\usepackage{epsfig}
\usepackage{graphicx}
\usepackage{amsmath}
\usepackage{amssymb}

\usepackage[pagebackref=true,breaklinks=true,letterpaper=true,colorlinks,bookmarks=false]{hyperref}

\iccvfinalcopy % *** Uncomment this line for the final submission

\def\iccvPaperID{3713} % *** Enter the ICCV Paper ID here

\ificcvfinal\fi

\begin{document}

%%%%%%%%% TITLE
\title{Learning Instance-level Spatial-Temporal Patterns for Person Re-identification}

\author{Min Ren\textsuperscript{1 2}\thanks{This work is done when Min Ren is an intern at JD AI Research.}, Lingxiao He\textsuperscript{3}, Xingyu Liao\textsuperscript{3},Wu Liu\textsuperscript{3}, Yunlong Wang\textsuperscript{2}, Tieniu Tan\textsuperscript{2}\\
\textsuperscript{1}University of Chinese Academy of Sciences\\
\textsuperscript{2}CRIPAC NLPR, Institute of Automation Chinese Academy of Sciences\\
\textsuperscript{3}JD AI Research\\
\{min.ren, yunlong.wang\}@cripac.ia.ac.cn, \{helingxiao3, liaoxingyu5, liuwu1\}@jd.com, tnt@nlpr.ia.ac.cn
%\and
}

\maketitle
\ificcvfinal\thispagestyle{empty}\fi

%%%%%%%%% ABSTRACT
\begin{abstract}
Person re-identification (Re-ID) aims to match pedestrians under dis-joint cameras.
Most Re-ID methods formulate it as visual representation learning and image search, and its accuracy is consequently affected greatly by the search space.
Spatial-temporal information has been proven to be efficient to filter irrelevant negative samples and significantly improve Re-ID accuracy.
However, existing spatial-temporal person Re-ID methods are still rough and do not exploit spatial-temporal information sufficiently.
In this paper, we propose a novel Instance-level and Spatial-Temporal Disentangled Re-ID method (InSTD), to improve Re-ID accuracy.
In our proposed framework, personalized information such as moving direction is explicitly considered to further narrow down the search space.
Besides, the spatial-temporal transferring probability is disentangled from joint distribution to marginal distribution, so that outliers can also be well modeled.
Abundant experimental analyses are presented, which demonstrates the superiority and provides more insights into our method.
The proposed method achieves mAP of~ 90.8\% on Market-1501 and 89.1\% on DukeMTMC-reID, improving from the baseline 82.2\% and 72.7\%, respectively. 
Besides, in order to provide a better benchmark for person re-identification, we release a cleaned data list of DukeMTMC-reID with this paper: \url{https://github.com/RenMin1991/cleaned-DukeMTMC-reID/}
\end{abstract}

%%%%%%%%% BODY TEXT
\section{Introduction}

Person re-identification aims to retrieve pedestrians across non-overlapping camera views. Most existing person re-identification methods focus on the visual feature representations of pedestrian images~\cite{2015An, 2018Learning, he2020fastreid, 2015Learning, 2016Person, Lin2017Deep, 2017Beyond, Chen2020Salience, 2016Deep}, such as appearance, clothes, and textures.
The auxiliary information of person images is also adopted recently, such as parsing information~\cite{ Song2018Mask,He2018CVPR, 2020Foreground, 2018MaskReID, Kalayeh2018Human}, pose of the pedestrians~\cite{Su2017Learning, 2019Pose, 2018Pose}, or human body key points~\cite{Wang2020High}. 
%However, these methods need expensive annotations to match the query and gallery images. And the computation complexity of these methods can be too high to be applied to large scale situations in practice.
However, the performances of these methods are still far from the requirements of real-world situations.
Because it is hard for visual representations to discriminate pedestrian with similar appearance and clothes.

\begin{figure}[t]
\begin{center}
   \includegraphics[width=\linewidth]{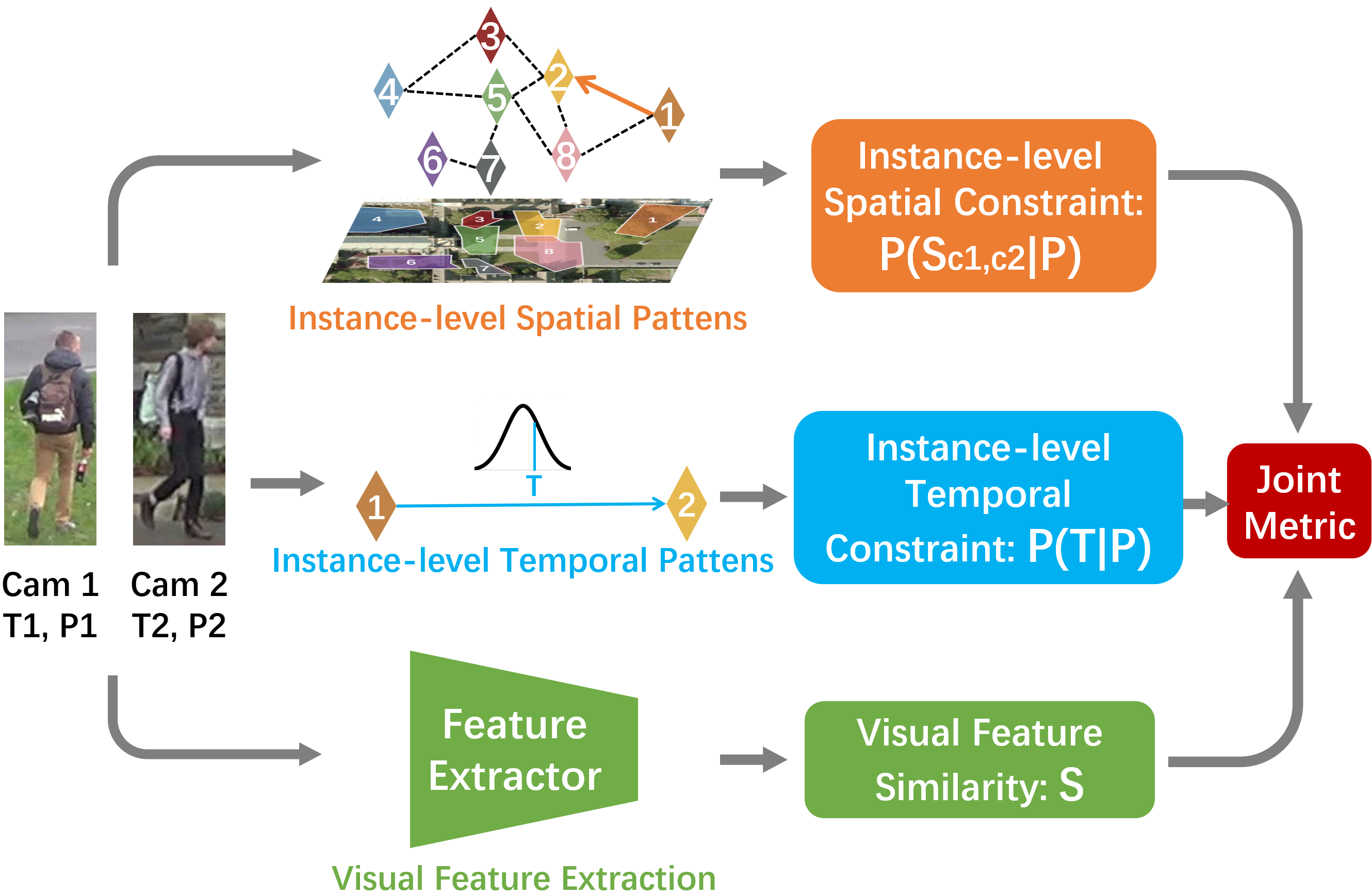}
\end{center}
   \caption{For each pair of pedestrian images, instance-level spatial and temporal constraints are provided separately by the proposed framework. Then they are adaptively combined with the visual feature similarity for matching.}
\label{fig:pipeline}
\vspace{-0.3cm}
\end{figure}

Recent methods model spatial-temporal patterns ~\cite{guangcong2019aaai, 2018Unsupervised, 2017Joint, 2016Camera} to filter out the irrelevant candidates and narrow down the search space.
Specifically, these methods mainly formulate spatial-temporal pattern as a joint distribution $P(S_{c_i,c_j},T)$, where $S_{c_i,c_j}$ means moving from \textit{camera i} to \textit{camera j}, $T$ means time interval.
It has been proven to be efficient to significantly improve re-identification accuracy.
However, there are two problems of the existing methods.
Firstly, the existing spatial-temporal methods only consider camera-level but neglect instance-level information.
The state information of each pedestrian is neglected while it is essential for spatial-temporal patterns of the person.
%
%For example, the spatial-temporal patterns of two persons who walk in opposite directions can be quite different. 
%
Secondly, existing methods formulate spatial-temporal patterns as a joint distribution, meaning that only those candidates matching both spatial and temporal priors can be matched. 
They are not robust to the outliers.% For example, a runner may not satisfy the temporal pattern but match the spatial pattern.

\iffalse
Some methods pay attention to the spatial-temporal constraints, which are implicit in the topology of cameras, to enhance the person re-identification performance~\cite{2016Camera, 2017Joint, 2018Unsupervised, guangcong2019aaai}. 
Compared with the structure information, the spatial-temporal information is much more economical but useful.
In these methods, hard or soft constraints are utilized to filter out the irrelevant gallery images so as to simplify the retrieval of pedestrians.
%
However, the existing spatial-temporal methods model the patterns according to the camera-level information, which may not provide reasonable constraints for each instance. 
The state information of each pedestrian are neglected while it is essential for spatial-temporal patterns of the person.
%For example, the proper spatial and temporal patterns of two persons who walk in opposite directions can be quite different, as shown in Fig.~\ref{fig:intro1}. 
Besides, the spatial and temporal patterns are coupled together in most of the current methods~\cite{2017Joint, 2018Unsupervised, guangcong2019aaai}. This coupling is somehow harmful to person re-identification. For example, a runner, who is moving faster than most pedestrians, is an outlier from the view of temporal patterns. But the person can be quite normal from terms of spatial transmission perspective. The existing methods are likely to filter out this runner because the spatial and temporal patterns are coupled together.
It can be worse in complex scenarios.
\fi

\begin{figure*}[t]
\begin{center}
   \includegraphics[width=0.87\linewidth]{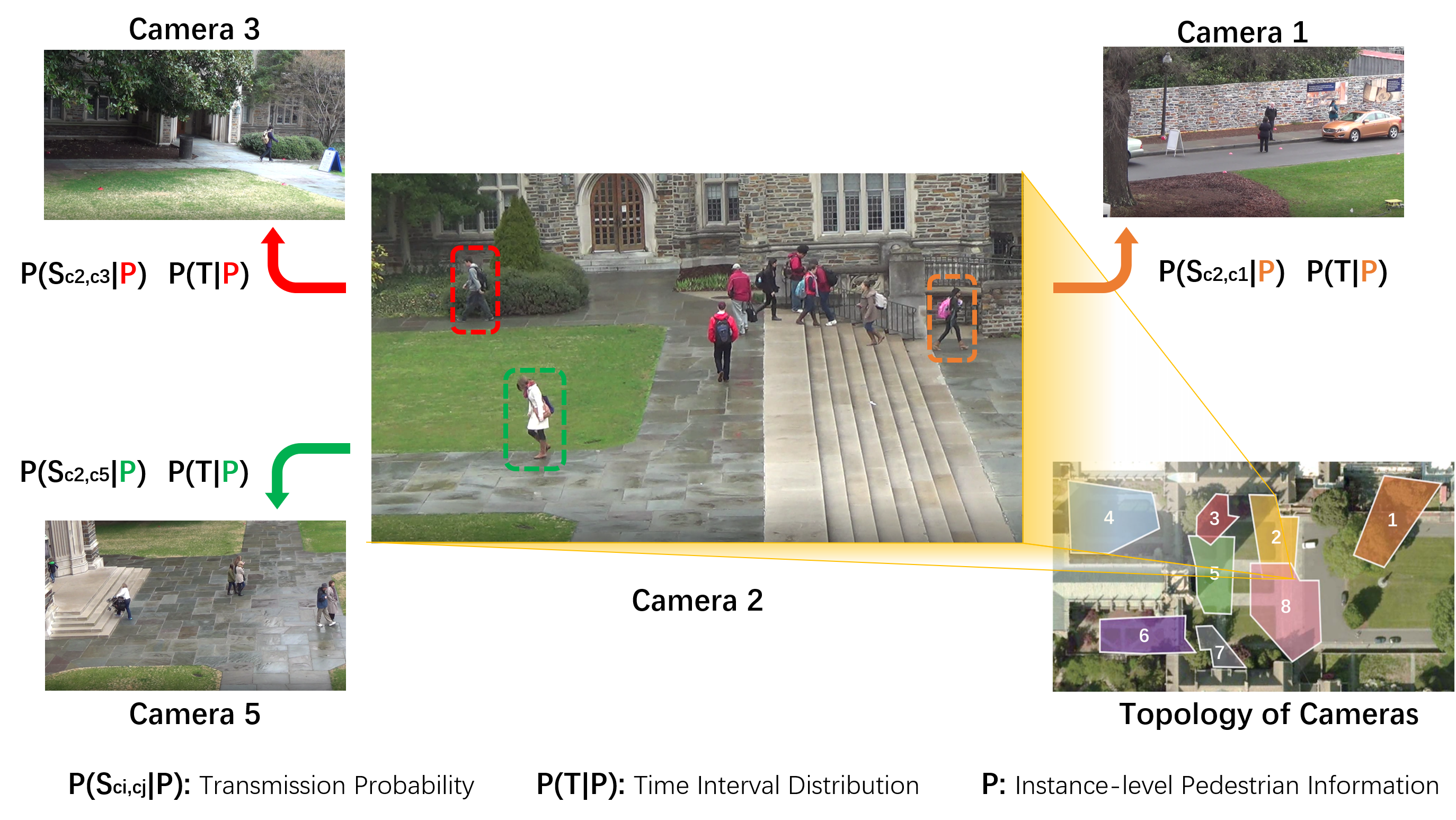}
\end{center}
\vspace{-0.3cm}
   \caption{Spatial-temporal patterns are implied in the topology of cameras. The spatial-temporal pattern between two cameras of a pedestrian are highly correlated with his/her moving direction. Pedestrians in the view of camera 2 may appear in camera 1, camera 3, or camera 5  after a certain time lapse. However, pedestrians with different states will appear in different cameras at different times. For example, the pedestrian in the red bounding box is much more likely to appear in Camera 3 than Camera 1, because he is moving towards the field of view of Camera 3. In the proposed method, the instance-level state information is adopted rather than modeling the spatial-temporal patterns on camera-level as the existing methods.}
\label{fig:intro1}
\vspace{-0.3cm}
\end{figure*}

To solve these problems, we propose a novel method named Instance-level and Spatial-Temporal Disentangled Re-ID (InSTD) to model the instance-level and spatial-temporal disentangled patterns.
Firstly, the traditional spatial-temporal pattern is updated to be conditional on instance-level state information.
Its formulation looks like $p(S_{c_i, c_j}, T|P)$, where $P$ is instance-level pedestrian information.
The walking direction of the pedestrian, which is the key instance-level state information, is taken into consideration in this paper.
The walking direction of a pedestrian is complimentary information of pedestrian detection and tracking.
It is useful because it is highly correlated with spatial-temporal patterns.
For example, a pedestrian, who is walking towards the west in the view of a camera, is more probable to appear in the view of the western cameras later, rather than the eastern cameras.
Meanwhile, it is economical because pedestrian detection and tracking are necessary steps before person re-identification in practice.
%

%There is no need to predict the walking direction by an extra model or man This key instance-level information can be 

Secondly, we disentangle the spatial-temporal pattern by constructing their marginal distribution, \ie transmission probability $P(S_{c_i,c_j}|P)$ and time interval distribution $P(T|P)$.
They are modeled separately and adaptively combined to handle outliers.
If the temporal (spatial) pattern of a pedestrian is unusual, the person may be normal in the term of spatial (temporal) patterns. The similarity metric should focus on the spatial (temporal) pattern.
For example, a runner, who is moving faster than most pedestrians, is an outlier from the view of temporal pattern. But the runner can be quite normal in terms of spatial transmission perspective.
It is harmful to model this runner by joint distribution of spatial and temporal patterns.
To this end, we propose a novel fusion approach to adaptively combine the spatial and temporal patterns.
The spatial patterns and temporal patterns are complementary, rather than in conflict as existing methods, so that outliers can also be well modeled.

The contributions of this paper can be summarized as follows:

\begin{itemize}

\item We present a novel instance-level method to model spatial-temporal patterns for person re-identification. The proposed method provides personalized predictions by leveraging the instance-level state information of each pedestrian.

\item The instance-level spatial-temporal patterns are decoupled into transmission probabilities and time interval distributions between cameras in the proposed method. The spatial and temporal patterns become complementary rather than in conflict as existing methods.

\item Without bells and whistles, the proposed method surpasses the baseline model based on visual features by 16.9\% on DukeMTMC-reID and 8.6\% on Market-1501 in the term of mAP, and outperforms the state-of-the-art method based on spatial-temporal patterns by 4.8\% on DukeMTMC-reID and 2.2\% on Market-1501.

\end{itemize}

%%%%%%%%%%%%%%%%%%%%%%%%%%%%%%%%%%%
%%%%%%%%%%%%%%%%%%%%%%%%%%%%%%%%%%%

\section{Related Work}
\label{sec:related_work}

%------------------------------------------------------------------------------
%------------------------------------------------------------------------------

%\subsection{Visual Features Based Re-identification}

% update by wangguanan
\subsection{Visual Features based Re-ID}

Person re-identification addresses the problem of matching pedestrian images across non-overlapping camera views~\cite{Gong2014Person}. Many studies exploit discriminative visual features~\cite{Apurva2011Multiple, Ma2018Covariance, Yang2014Salient}. 
%Ma \etal ~\cite{Ma2018Covariance} propose a person image representation which relies on the combination of Biologically Inspired Features (BIF) and Covariance descriptors. 
%A salient color names based color descriptor (SCNCD)~\cite{Yang2014Salient} is proposed for person image matching. 
%Apurva \etal~\cite{Apurva2011Multiple} present a part-based spatio-temporal model that learns a person's characteristic appearance. 
%Metric learning methods are also introduced into the field of person re-identification~\cite{Martin2012Large, Zheng2013Reidentification}. 
%Zheng \etal~\cite{Zheng2013Reidentification} formulate person reidentification as-a relative distance comparison (RDC) learning problem to learn the optimal similarity measure between a pair of person images. 
%Martin \etal~\cite{Martin2012Large} introduce a simple though effective strategy to learn a distance metric from equivalence constraints.

Deep learning algorithms foster significant improvements in the field of person re-identification.
Some researchers attempt to explore effective convolutional neural networks~\cite{Hermans2017In, 2015An, Wang2017P2SNet, Wang2016DARI, Shen2018PersonRW, Ding2015Deep, Lin2017Deep, Chen2020Salience, Zhang2020Relation, Zheng2018Person}. 
%Ejaz \etal~\cite{2015An} propose a deep learning architecture for simultaneously learning features and a corresponding similarity metric for person re-identification. Wu \etal~\cite{Lin2017Deep} introduce a hybrid deep architecture that combines Fisher vectors and deep neural networks. Deep graph models are adopted by Shen \etal~\cite{Shen2018PersonRW}.
%Chen \etal~\cite{Chen2020Salience} propose a Salient Feature Extraction (SFE) unit to suppress the salient features and adaptively extract other potential salient features to obtain different clues of pedestrians.
%Zhang \etal~\cite{Zhang2020Relation} try to capture the global structural information for relation-aware global attention.
Some studies explore training strategies and loss functions for person re-identification~\cite{Hermans2017In, Wang2017P2SNet, Wang2016DARI, Ding2015Deep}. 
Recently, some studies leverage the structure information of person images, such as parsing information~\cite{Song2018Mask, 2020Foreground, 2018MaskReID, Kalayeh2018Human}, pose of the pedestrians~\cite{Su2017Learning, 2019Pose, 2018Pose}, or human body key points~\cite{Wang2020High}.

%Li \etal~\cite{Li2017Learning} propose a multi-scale context-aware network (MSCAN) to learn powerful features over the full body and body parts.
%Zhao \etal~\cite{Zhao2017Spindle} propose a convolutional neural network based on human body region guided multi-stage feature decomposition and tree-structured competitive feature fusion.
%A pose-driven deep learning model is proposed by Su \etal~\cite{Su2017Learning} to alleviate the pose variations and learn robust feature representations from both the global image and different local parts.
%The segmentation masks of person images are integrated into person re-identification by recent work~\cite{Kalayeh2018Human, Song2018Mask}.
%Gao \etal~\cite{Gao2020Pose} propose a pose-guided attention method for part feature pooling that exploits more discriminative local features.
%Wang \etal~\cite{Wang2020High} propose a framework by learning high-order relation and topology information for discriminative features and robust alignment.

However, appearance-based methods are still far from practical applications.
They are not discriminative enough in complex scenarios where pedestrians may exhibit similar appearance and clothes.
It is hard to further improve the performance using only appearance-based features.

%------------------------------------------------------------------------------
%------------------------------------------------------------------------------

%\subsection{Spatial-temporal Based Person Re-identification}

% update by wangguanan
\subsection{Spatial-temporal Person Re-ID}

There are some researchers who have paid attention to the topology of cameras since the spatial-temporal patterns implied in the topology are essential for cross-camera retrieval. 
The spatial-temporal constraints are utilized to filter out the irrelevant gallery images~\cite{guangcong2019aaai, 2018Unsupervised, 2017Joint, 2016Camera}. 
Huang \etal~\cite{2016Camera} propose a method to take both visual feature representation and spatial-temporal constraints into consideration for person re-identification. However, this method makes a strong assumption that the time intervals between cameras follow Weibull distribution. This assumption is invalid for complex scenarios.  
Cho \etal ~\cite{2017Joint} propose a framework to integrate camera network topology into person re-identification. However, the temporal constraints are simply realized by time thresholds, which cannot handle massive gallery images and complex cases in practice.
Lv \etal~\cite{2018Unsupervised} propose a method that leverage the spatial-temporal constraints for cross-dataset person re-identification. The spatial-temporal constraints improve the performance on the target dataset by enhancing the pseudo label during training. 
It is not proper to be directly applied to general person re-identification tasks.
Wang \etal~\cite{guangcong2019aaai} propose a two-stream architecture to apply spatial-temporal constraints to person re-identification. However, the spatial-temporal constraints are coupled together in this method, which is harmful to recalling positive samples. 
All these spatial-temporal person Re-ID methods establish the patterns based on the camera-level information, which means they can not provide fine-grained constraints for each instance.

Different from existing spatial-temporal methods, our method models spatial-temporal patterns at the instance-level to filter out more irrelevant gallery images and provide personalized predictions. And the spatial-temporal patterns are decoupled into transmission probabilities and time interval distributions to make them mutually beneficial rather than in conflict.

%%%%%%%%%%%%%%%%%%%%%%%%%%%%%%%%%%%
%%%%%%%%%%%%%%%%%%%%%%%%%%%%%%%%%%%

\section{Method}

The instance-level spatial constraint \ie transmission probabilities, and temporal constraint \ie time interval distributions are detailed separately in this section.
Then the adaptive combined metric is presented.

%------------------------------------------------------------------------------
%------------------------------------------------------------------------------

\begin{figure}[t]
\begin{center}
   \includegraphics[width=\linewidth]{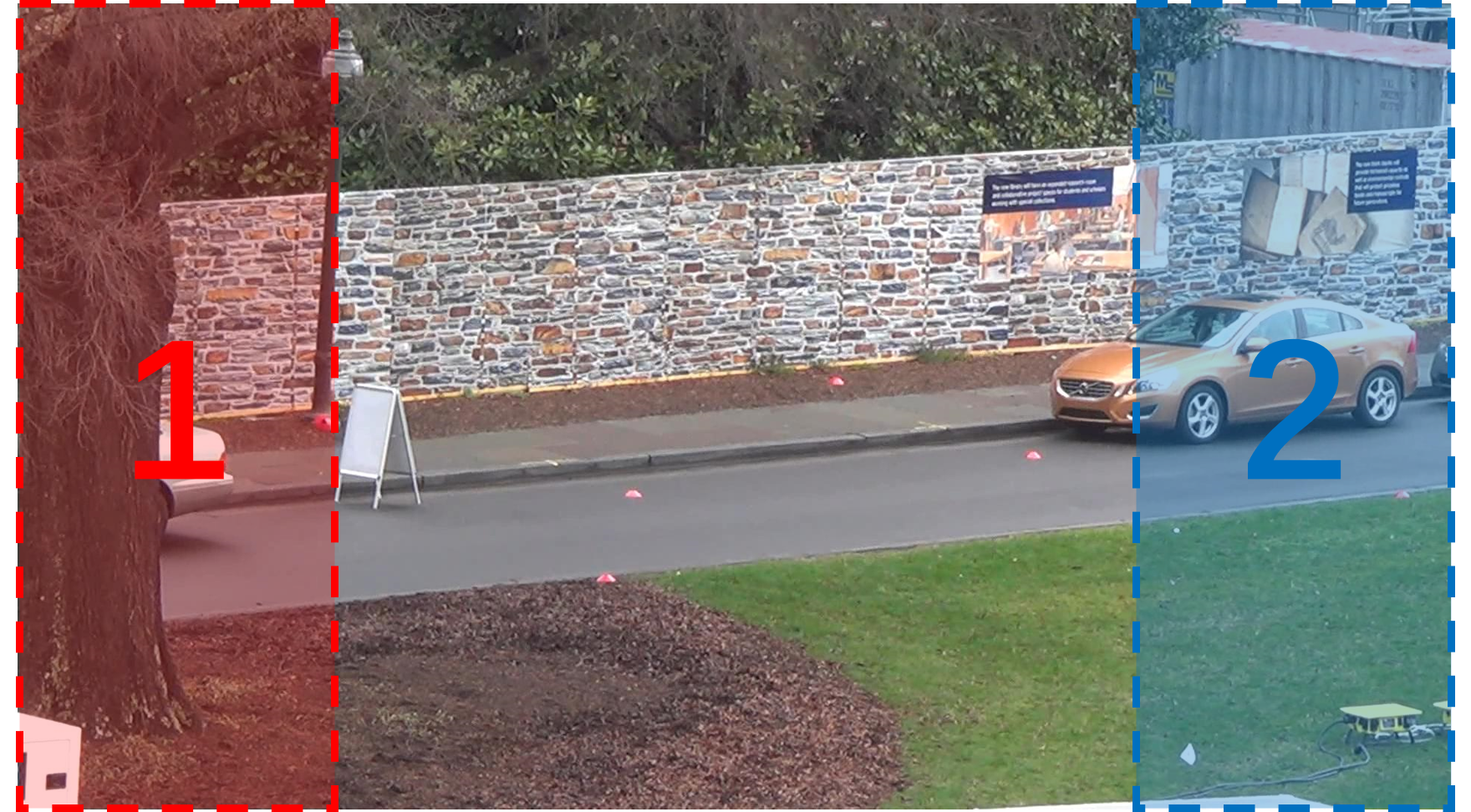}
\end{center}
   \caption {View of the first camera of DukeMTMC-reID. The state set of this camera contains two states: walking towards the red zone and walking towards the blue zone.}
\label{fig:cam1}
\vspace{-0.3cm}
\end{figure}

\subsection{Instance-level Spatial Constraint}

The spatial constraint between two cameras is described by the transmission probability of the cameras, which means how tightly the two cameras correlate. Formally, we model the transmission probability by a conditional probability:
\begin{equation}
p_{i,j} = Pr(C_{l}=j|C_{e}=i)
\label{equ:spatial1}
\end{equation}
where $i$ and $j$ are the indexes of cameras, $C_{e}$ is the camera that a person appears earlier, $C_{l}$ is the camera that the same person appears later.
It is the probability that a person appears in the view of camera $j$ later on the condition that this person has appeared in the view of camera $i$.
The conditional probability in Eq.~\ref{equ:spatial1} can be easily calculated:
\begin{equation}
p_{i,j} = \frac{Pr(C_{l}=j,C_{e}=i)}{Pr(C_{e}=i)}
\label{equ:spatial2}
\end{equation}

The higher the conditional probability means the person in the view of camera $i$ is more likely to appear in the view of camera $j$ later. The time interval between camera $i$ and $j$ is not involved here. Note that $p_{i,j} \neq p_{j,i}$ in most cases.

However, the spatial patterns of persons appear in the same camera can be different, as shown in Fig.~\ref{fig:intro1} . To address this problem, we introduce the instance-level state information of a person into the conditional probability:
\begin{equation}
p^s_{i,j} = Pr(C_{l}=j|C_{e}=i, s_{e}=s)
\label{equ:spatial_state}
\end{equation}
where $s_{e}$ is the state of a person in the view of $C_{e}$, $s\in S_i$, $S_i$ is the set of states: 
\begin{equation}
S_i = \{ s_1, s_2, ..., s_{n_i}\}
\label{equ:state_set}
\end{equation}
where $n_i$ is the number of states of camera $i$.

The instance-level states are represented by walking directions of pedestrians. 
For example, the view of the first camera of DukeMTMC-reID~\cite{Ergys2016Performance} is shown in Fig.~\ref{fig:cam1}. The state set of this camera contains two states: walking towards the red zone and walking towards the blue zone. The state sets of the rest cameras are defined similarly, and the illustrations of other cameras can be found in the supplementary material.

Hence, the instance-level transmission probability can be calculated: 
\begin{equation}
p^s_{i,j} = \frac{Pr(C_{l}=j,C_{e}=i, s_{e}=s)}{Pr(C_{e}=i, s_{e}=s)}
\label{equ:spatial3}
\end{equation}

%------------------------------------------------------------------------------
%------------------------------------------------------------------------------

\subsection{Instance-level Temporal Constraint}

The temporal constraint is described by the time interval distribution, which represents the time lapse for a pedestrian to transfer between two cameras. Formally, we model the time interval distribution by a conditional probability density function:
\begin{equation}
f_{i,j}(\delta) = \frac{\mathrm{d}F_{i,j}(\delta)}{\mathrm{d}\delta}
\label{equ:temporal1}
\end{equation}
where $\delta$ is the transfer time,  $F_{i,j}(\delta)$ is the cumulative distribution function, which is a conditional probability:
\begin{equation}
F_{i,j}(\delta) = Pr(\Delta \leq \delta | C_{e}=i, C_{l}=j)
\label{equ:temporal2}
\end{equation}

It can be harmful to recalling positive samples that fitting $f_{i,j}(\delta)$ or $F_{i,j}(\delta)$ into a closed-form probability distribution.
Hence, a non-parameter estimation method is adopted in our method. Specially, we use Parzen window with Gaussian kernel to estimate $f_{i,j}(\delta)$:
\begin{equation}
f_{i,j}(\delta) = \frac{1}{Z_{i,j}} \sum_n H_{i,j}(\delta) K(n-\delta) 
\label{equ:parzen1}
\end{equation}

\begin{equation}
H_{i,j}(\delta)=\left\{
\begin{array}{cl}
1 & \delta \in \mathcal{D}_{i,j} \\
0 & otherwise
\end{array}
\right.
\label{equ:parzen2}
\end{equation}
where $K(\cdot)$ is the kernel function, ${Z_{i,j}} = \sum_\delta H_{i,j}(\delta)$ is a normalized factor, $\mathcal{D}_{i,j}$ is the time interval set between camera $i$ and camera $j$ of training samples.
%Note that the time interval distributions are normalized separately, which means the time interval distributions are not involved with the transmission probabilities.

However, the time interval distribution of persons appear in the same camera can be different, as we have mentioned before. Similar to the instance-level transmission probabilities, the instance-level state information of a person is introduced into the conditional probability to address this problem:
\begin{equation}
f^s_{i,j}(\delta) = \frac{\mathrm{d}F^s_{i,j}(\delta)}{\mathrm{d}\delta}
\label{equ:temporal3}
\end{equation}
\begin{equation}
F^s_{i,j}(\delta) = Pr(\Delta \leq \delta | C_{e}=i, C_{l}=j, s_{e}=s)
\label{equ:temporal4}
\end{equation}
where $s_{e}$ is the instance-level state of a person in the view of $C_{e}$. The moving direction is also considered as the key state information.
Similar with Eq.~\ref{equ:parzen1} , instance-level time interval distribution $f^s_{i,j}(\delta)$ can be estimated:
\begin{equation}
f^s_{i,j}(\delta) = \frac{1}{Z^s_{i,j}} \sum_n H^s_{i,j}(\delta) K(n-\delta) 
\label{equ:parzen3}
\end{equation}

\begin{equation}
H^s_{i,j}(\delta)=\left\{
\begin{array}{cl}
1 & \delta \in \mathcal{D}^s_{i,j} \\
0 & otherwise
\end{array}
\right.
\label{equ:parzen4}
\end{equation}
where the normalized factor ${Z^s_{i,j}} = \sum_\delta H^s_{i,j}(\delta)$, $\mathcal{D}^s_{i,j}$ is a subset of $\mathcal{D}_{i,j}$, it contains the samples subject to $s_e=s$.

%------------------------------------------------------------------------------
%------------------------------------------------------------------------------

\subsection{Joint Metric}

Given the transmission probability and time interval distribution affiliated to instance-level state information, the spatial-temporal probability is the fusion of them:
\begin{equation}
\mathcal{P} = \mathcal{F}(p_{spa}, p_{tem})
\label{equ:fusion1}
\end{equation}
where $p_{spa} = p^s_{i,j}$ is the instance-level transmission probability of two images in Eq.~\ref{equ:spatial3}, and $p_{tem} = f^s_{i,j}(\delta)$ is the instance-level time interval probability in Eq.~\ref{equ:temporal3}. And the final joint metric of two images is:
\begin{equation}
\mathcal{S}\cdot \mathcal{P} = \mathcal{S}\cdot \mathcal{F}(p_{spa}, p_{tem})
\label{equ:fusion2}
\end{equation}
where $\mathcal{S}$ is the visual feature similarity.

A straightforward way to fuse both components is multiplying $p_{spa}$ and $p_{tem}$ together. 
However, the constraint realized by directly multiplying is too strict for person re-identification.
For example, the spatial constraint of a gallery image given by transmission probability is 0.9, the temporal constraint given by time interval distribution is 0.01; the spatial constraint of another gallery image is 0.1, the temporal constraint is 0.1. 
The first gallery image should be ranked higher than the second one because their temporal constraints are similar actually, while the spatial constraints have a significant gap.
However, if we fuse the spatial constraint and temporal constraint by multiplying, the second image will be ranked higher instead.
If a spatial/temporal pattern of a pedestrian is unusual, the person may be normal in terms of temporal/spatial patterns. This kind of samples should not be removed recklessly.
Hence, the spatial-temporal probability $\mathcal{F}(p_{spa}, p_{tem})$ should be fairly high when only one of them is high. Fusion by multiplying directly is not proper here obviously.

The spatial-temporal probability in our method is defined as:
\begin{equation}
\mathcal{P} = \frac{1}{1+e^{-(\alpha p_{spa} + \beta p_{tem})}}
\label{equ:fusion3}
\end{equation}
where $\alpha$ and $\beta$ are scaling parameters of similarity fusion. The spatial and temporal constraints are adjusted by $\alpha$ and $\beta$ separately.

The spatial-temporal factor is scaled into $[0.5, 1)$. The constraint is relaxed properly when the spatial-temporal probability is low. And the value of $\mathcal{P}$ stays stable when $p_{spa}$ or $p_{tem}$ is low.

%------------------------------------------------------------------------------
%------------------------------------------------------------------------------

\subsection{Implementation Details}

More details are presented in this subsection.
The moving direction of a pedestrian is complimentary information of pedestrian detection and tracking, which is a necessary step before person re-identification in practice.
There is no need to predict the moving direction by an extra model or manual annotations.
In our experiments, the moving directions of samples in the field of one camera are confirmed by tracking them in the original video of this camera.
Actually, the moving direction of a pedestrian can be confirmed within five consecutive frames in most cases, which can be easily derived from existing tracking methods.

A pretrained ResNet-50 is adopted as baseline for feature extraction.
We set the standard deviation of the Gaussian kernel for distribution estimation to 100.
As for the scaling parameters, $\alpha$, $\beta$ in Eq.~\ref{equ:fusion3} are set to 0.15 and 1 respectively.

%%%%%%%%%%%%%%%%%%%%%%%%%%%%%%%%%%%
%%%%%%%%%%%%%%%%%%%%%%%%%%%%%%%%%%%

\section{Experiments}

In this section, we evaluate our method on two large scale person re-identification benchmark datasets, \ie Market-1501~\cite{Zheng2015Scalable} and DukeMTMC-reID~\cite{Ergys2016Performance}. Then, more experimental analysis is presented.

%------------------------------------------------------------------------------
%------------------------------------------------------------------------------

\subsection{Datasets and Evaluation Protocol}

Market-1501 dataset~\cite{Zheng2015Scalable} is collected at a university campus. A total of six cameras are used, including 5 high-resolution cameras, and one low-resolution camera. It contains 32,668 annotated bounding boxes of 1,501 identifies, plus a distractor set of over 500K images. The pedestrians are detected by  Deformable Part Model (DPM). Among them, the training set consists of 12,936 images from 751 identities, the gallery set contains 19,732 images from other 750 identities and all the distractors. 3,368 hand-drawn bounding boxes from 750 identities are used as the query images. In this dataset, each image contains its camera index and time stamp.

DukeMTMC-reID~\cite{Ergys2016Performance} is a subset of DukeMTMC dataset for image-based person re-identification. There are eight cameras in total. 1,404 identities appear in more than one camera and 408 identities (distractor) appear in only one camera. 702 identities are used for training, and the other 702 identities plus distractors are used for testing. One image for each identity in each camera is picked as a query, and the other images are put into the gallery. Each image contains its camera index and time stamp.

We use two performance indexes as in most person re-identification literature.
The first is mean average precision (mAP). The average precision (AP) of a query is the area under the Precision-Recall curve, which means both precision and recall rate is taken into consideration. Hence, the mean average precision among all query images is a comprehensive performance index for person re-identification.
The second is the cumulative matching characteristic (CMC) \ie the top-k accuracy. Hence, the cumulative matching characteristic emphasizes precision rather than recall rate.

%------------------------------------------------------------------------------
%------------------------------------------------------------------------------

\subsection{Comparisons to the State-of-the-Art}

The proposed method is compared with fourteen existing state-of-the-art methods, which can be categorized into four groups.
The first group of methods extract visual features directly from the person images, including PCB~\cite{2017Beyond}, VPM~\cite{Perceive2020Perceive}, and BOT~\cite{Luo2019Bag}. These methods explore various aspects of visual feature extraction, including the structure of convolutional neural networks, training strategy, data augmentation, and loss function.
The second group of methods adopt human parsing information for person re-identification, including SPReID~\cite{Kalayeh2018Human}, MGCAM~\cite{Song2018Mask}, MaskReID~\cite{2018MaskReID} and FPR~\cite{2020Foreground}. These methods introduce semantic information to the person images for better image alignment and feature matching.
The third group of methods leverage the pose or key points of person images, including PDC~\cite{Su2017Learning}, Pose-transfer~\cite{2018Pose}, PSE~\cite{2017A}, PGFA~\cite{2019Pose} and HOReID~\cite{Wang2020High}. These methods attempt to overcome the various human pose by taking the pose information or key points of the human body into consideration.
The fourth group of methods utilize the spatial-temporal information to enhance the person re-identification, including TFusion-sup~\cite{2018Unsupervised} and st-ReID~\cite{guangcong2019aaai}. These methods use hard or soft constraints to narrow the number of gallery images.

\begin{table}[t]
\begin{center}
\setlength{\tabcolsep}{1mm}
{
\begin{tabular}{l | c | c | c | c}
\hline
\bf{Methods} & \bf{mAP} & \bf{Rank-1} & \bf{Rank-5} & \bf{Rank-10}  \\
\hline
PCB~\cite{2017Beyond} & 77.4\% & 92.3\% & 97.2\% & 98.2\% \\
VPM~\cite{Perceive2020Perceive} & 80.8\% & 93.0\% & 97.8\% & 98.8\% \\
BOT~\cite{Luo2019Bag} & 85.9\% & 94.5\% & - & - \\

\hline
SPReID~\cite{Kalayeh2018Human} & 81.3\% & 92.5\% & 97.15\% & 98.1\% \\ 
MGCAM~\cite{Song2018Mask} & 74.3\% & 83.8\% & - & - \\
MaskReID~\cite{2018MaskReID} &75.4\% & 90.4\% & - & - \\
FPR~\cite{2020Foreground} & 86.6\% & 95.4\% & - & - \\

\hline
PDC~\cite{Su2017Learning} & 63.4\% & 84.1\% & - & - \\
Pose-transfer~\cite{2018Pose} & 68.9\% & 87.7\% & - & - \\
PSE~\cite{2017A} & 69.0\% & 87.7\% & 94.5\% & 96.8\% \\
PGFA~\cite{2019Pose} & 76.8\% & 91.2\% & - & - \\
HOReID~\cite{Wang2020High} & 84.9\% & 94.2\% & - & - \\

\hline
Baseline & 82.2\% & 93.6\% & 98.4\% & 99.0\% \\
%TFusion-sup~\cite{2018Unsupervised} & - & 73.1\% & 86.4\% & 90.5\% \\
Baseline+st-ReID~\cite{guangcong2019aaai} & 88.6\% & 96.9\% & 99.2\% & 99.5\% \\

%\hline

\textbf{Baseline+InSTD} & \bf{90.8\%} & \bf{97.6\%}  & \bf{99.5\%} & \bf{99.7\%} \\
\hline
\end{tabular}}
\end{center}
\caption{Comparison with the state-of-the-arts methods on Market-1501. Group 1: vanilla deep learning based methods. Group 2: human-parsing information based methods. Group 3: pose or key points based methods. Group 4: spatial-temporal methods.}
\label{tab:market}
\vspace{-0.3cm}
\end{table}

The experiment results on Market-1501 are shown in Tab.~\ref{tab:market}, and the results on DukeMTMC-reID are shown in Tab.~\ref{tab:duke}.
Our method outperforms all of the existing methods on both datasets. Comparing to the baseline model, which is a ResNet-50, our method improves the mAP by 8.6\% on Market-1501 and 16.5\% on DukeMTMC-reID, improve the Rank-1 accuracy by 4\% on Market-1501 and 10\% on DukeMTMC-reID. Our method achieves significant improvements especially in terms of mAP.

Comparing to the methods in the first three groups, the advantage of our method is obvious. Besides,  the compared methods in the second and third groups need expensive annotations, such as key points, pixel-wise parsing maps, and masks, to match the query and gallery images. Our method adopts economical information \ie camera ID, timestamp, and state information.

The disadvantages of the methods in the fourth group have been interpreted in Sec.~\ref{sec:related_work}. And the interpretations have been demonstrated by the results of experiments in this subsection.
Given the same baseline model, our method outperforms the st-ReID by a remarkable margin, especially in terms of mAP (2.2\% on Market-1501 and 4.8\% on DukeMTMC-reID). These results indicate that our method has evident advantages over existing spatial-temporal methods.

\begin{table}[t]
\begin{center}
\setlength{\tabcolsep}{1mm}
{
\begin{tabular}{l | c | c | c | c}
\hline
\bf{Methods} & \bf{mAP} & \bf{Rank-1} & \bf{Rank-5} & \bf{Rank-10}  \\
\hline

PCB~\cite{2017Beyond} & 66.1\% & 81.7\% & 89.7\% & 91.9\% \\
VPM~\cite{Perceive2020Perceive} & 72.6\% & 83.6\% & 91.7\% & 94.2\% \\
BOT~\cite{Luo2019Bag} & 76.4\% & 86.4\% & - & - \\

\hline
SPReID~\cite{Kalayeh2018Human} & 70.9\% & 84.4\% & 91.8\% & 93.7\% \\ 
MGCAM~\cite{Song2018Mask} & 46.0\% & 46.7\% & - & - \\
MaskReID~\cite{2018MaskReID} &61.89\% & 78.86\% & - & - \\
FPR~\cite{2020Foreground} & 78.4\% & 88.6\% & - & - \\

\hline
Pose-transfer~\cite{2018Pose} & 56.9\% & 78.5\% & - & - \\
PSE~\cite{2017A} & 62.0\% & 79.8\% & 89.7\% & 92.2\% \\
PGFA~\cite{2019Pose} & 65.5\% & 82.6\% & - & - \\
HOReID~\cite{Wang2020High} & 75.6\% & 86.9\% & - & - \\

\hline
Baseline & 72.7\% & 85.7\% & 90.9\% & 93.5\% \\
Baseline+st-ReID~\cite{guangcong2019aaai} & 84.3\% & 94.1\% & 96.3\% &97.2\% \\

%\hline

\textbf{Baseline+InSTD} & \bf{89.1\%} & \bf{95.7\%}  & \bf{97.2\%} & \bf{98.0\%} \\
\hline
\end{tabular}}
\end{center}
\caption{Comparison with state-of-the-arts for person re-identification on DukeMTMC-reID~\cite{Ergys2016Performance}. Group 1: vanilla deep learning based methods. Group 2: human-parsing information based methods. Group 3: pose or key points based methods. Group 4: spatial-temporal methods.}
\label{tab:duke}
\vspace{-0.3cm}
\end{table}

\begin{figure}[t]
\begin{center}
   \includegraphics[width=\linewidth]{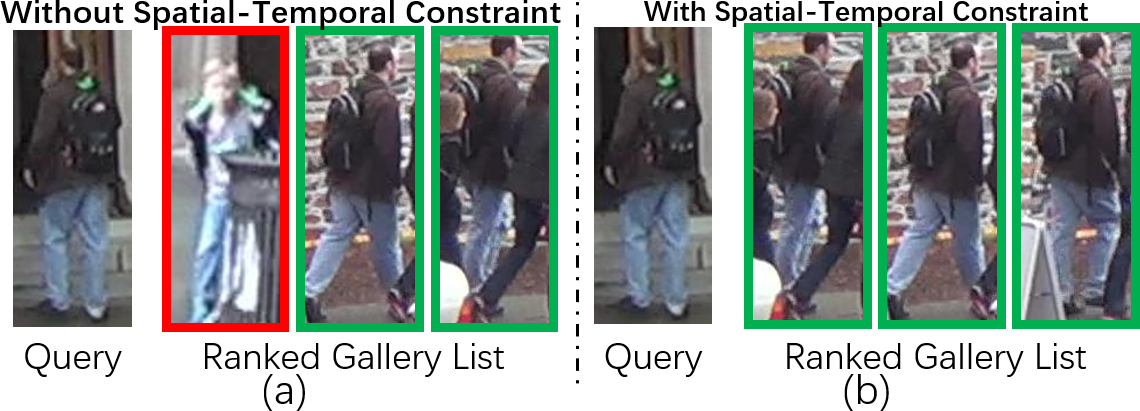}
\end{center}
\vspace{-0.3cm}
   \caption {\textbf{(a)}: The appearance of the pedestrian in the red bounding box, who is mistakenly ranked first, is similar to the query image. The visual representation cannot distinguish it from the correct identifies. \textbf{(b)}: The incorrect pedestrian is filtered out by the spatial-temporal constraint.}
\label{fig:suc1}
%\vspace{-0.1cm}
\end{figure}

To show the effect of spatial-temporal constraint, an example from DukeMTMC-reID is presented in Fig.~\ref{fig:suc1}.
The appearance of the pedestrian in the red bounding box, who is mistakenly ranked first, is similar to the query image. The visual representation cannot distinguish it from the correct identifies as shown in Fig.~\ref{fig:suc1} (a).
The incorrect pedestrian, which is difficult to discriminate for the visual representation, is filtered out by the spatial-temporal constraint as shown in Fig.~\ref{fig:suc1} (b).

\begin{figure}[t]
\begin{center}
   \includegraphics[width=\linewidth]{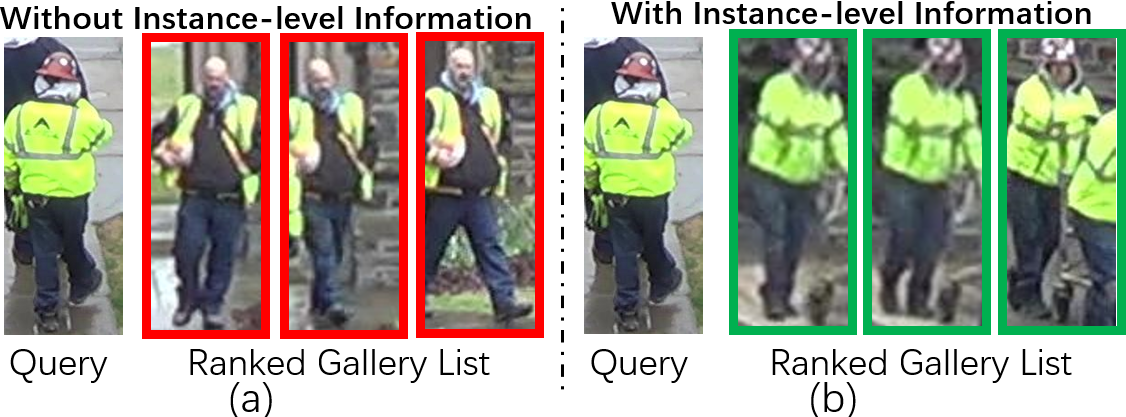}
\end{center}
\vspace{-0.3cm}
   \caption {\textbf{(a)}: The top three of the ranked list are wrong samples because the spatial-temporal constraints are misguided without instance-level information. \textbf{(b)}: The spatial-temporal constraints are more reliable because of the instance-level state.}
\label{fig:suc2}
\vspace{-0.3cm}
\end{figure}

The effect of instance-level information are shown in Fig.~\ref{fig:suc2}.
The spatial-temporal constrains may be misguided in complex scenarios, as shown in Fig.~\ref{fig:suc2} (a).
The instance-level information can make the spatial-temporal constrains more reliable for person re-identification.
The incorrect pedestrians are filtered out by the instance-level state as shown in Fig.~\ref{fig:suc2} (b).

%%%%%%%%%%%%%%%%%------------------------------------
%%%%%%%%%%%%%%%%%------------------------------------

\subsection{Experiments on Different Feature Extractors}
\vspace{-0.15cm}
The proposed method can be applied to different feature extractors.
To verify its effectiveness, we evaluate the proposed method based on other two feature extractors: PCB~\cite{2017Beyond}, and VPM~\cite{Perceive2020Perceive}.

The results are shown in Tab.~\ref{tab:diff_feat}.
Our method consistently improves the performance of all feature extractors.
Our method gains significant 20\%/11\% improvement in mAP/rank-1 accuracy for PCB~\cite{2017Beyond}, and 16\%+/10\%+ improvement for the other two feature extractors.
Comparing to st-ReID~\cite{guangcong2019aaai}, which is also based on spatial-temporal constraints, our method achieves consistent improvements too.
Our method outperforms st-ReID~\cite{guangcong2019aaai} by 4\%+/0.6\%+ improvement in mAP/rank-1 accuracy for all of the feature extractors.

The results show that our method can be generalized to different feature extractors.
Moreover, the results demonstrate the advantages of our method comparing to the existing spatial-temporal based method.

\begin{table}[t]
\begin{center}
\setlength{\tabcolsep}{0.5mm}
{
\begin{tabular}{l | c | c | c | c}
\hline
\bf{Methods} & \bf{mAP} & \bf{Rank-1} & \bf{Rank-5} & \bf{Rank-10}  \\
\hline

PCB~\cite{2017Beyond} & 66.1\% & 81.7\% & 89.7\% & 91.9\% \\
PCB~\cite{2017Beyond}+st-ReID~\cite{guangcong2019aaai} & 80.9\% & 92.1\% & 95.4\% & 96.6\% \\
\textbf{PCB~\cite{2017Beyond}+InSTD} & 86.1\% & 92.7\% & 96.5\% & 97.6\% \\

\hline

VPM~\cite{Perceive2020Perceive} & 72.6\% & 83.6\% & 91.7\% & 94.2\% \\
VPM~\cite{Perceive2020Perceive}+st-ReID~\cite{guangcong2019aaai} & 84.9\% & 94.2\% & 96.1\% & 96.9\% \\
\textbf{VPM~\cite{Perceive2020Perceive}+InSTD} & 89.3\% & 95.1\% & 97.0\% & 97.9\% \\

%\hline

%Baseline & 72.7\% & 85.7\% & 90.9\% & 93.5\% \\
%Baseline+st-ReID~\cite{guangcong2019aaai} & 84.3\% & 94.1\%  & 96.3\% & 97.2\% \\
%\textbf{Baseline+InSTD} & 89.1\% & 95.7\%  & 97.2\% & 98.0\% \\

\hline
\end{tabular}}
\end{center}
\caption{Effects on different feature extractors. The experiments are conducted on DukeMTMC-reID~\cite{Ergys2016Performance}}
\label{tab:diff_feat}
\end{table}

%--------------------------------------------------------------------------

\subsection{Analysis of Scaling Parameters}
\begin{figure}[t]
\begin{center}
   \includegraphics[width=\linewidth]{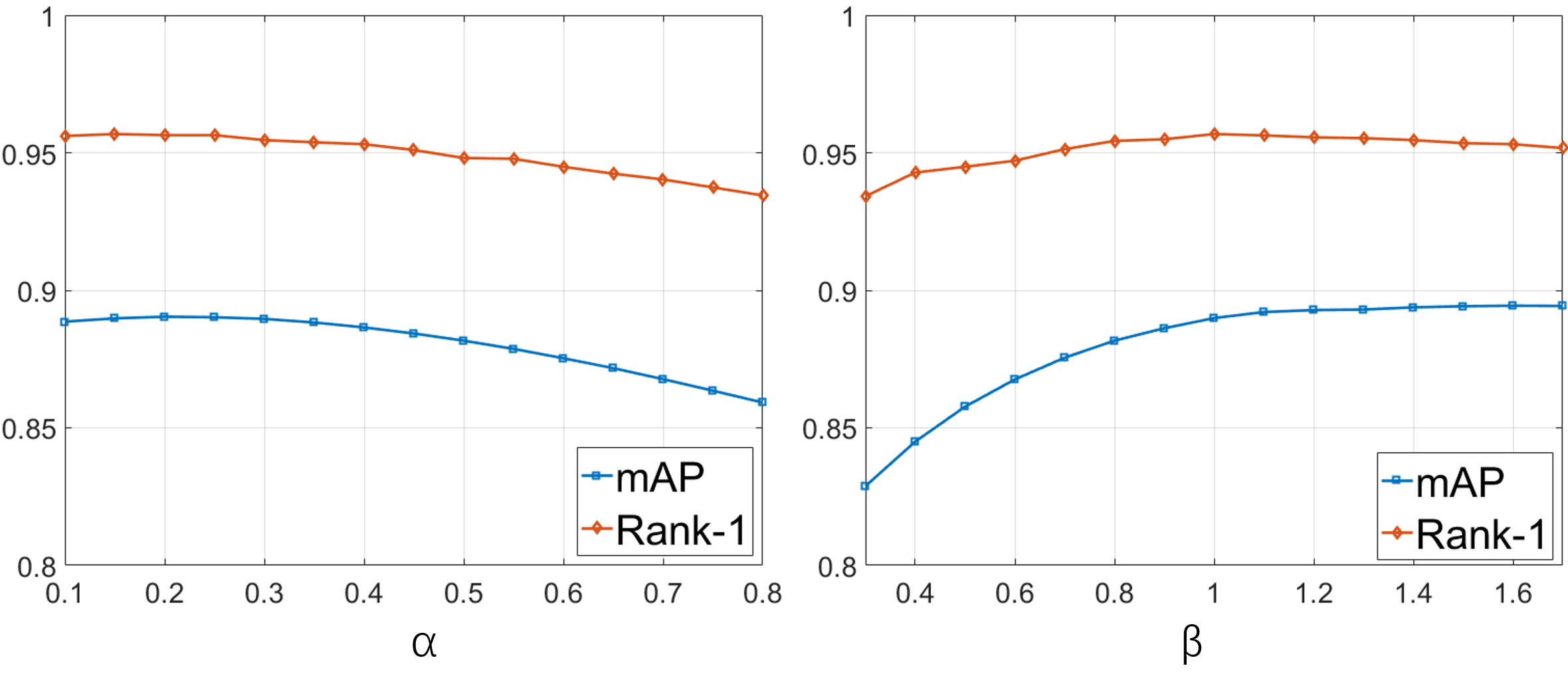}
\end{center}
\vspace{-0.4cm}
   \caption {Result of sensitivity analysis experiments on $\alpha$ and $\beta$ in Eq.~\ref{equ:fusion3}. When analyzing one of them, the other one is fixed as its optimal value. The experimental results show that our method is is insensitive to fusion parameters.}
\label{fig:param}
\vspace{-0.3cm}
\end{figure}

\vspace{-0.15cm}
To investigate the impact of two scaling parameters, $\alpha$ and $\beta$ in Eq.~\ref{equ:fusion3}, we conduct two sensitivity analysis experiments on $\alpha$ and $\beta$.
The results are shown in Fig.~\ref{fig:param}.
When analyzing one of them, the other one is fixed as its optimal value: $\alpha=0.15$, $\beta=1$.
As we can observe, our method nearly keeps the best performance when $\alpha$ is in the range of 0.1 to 0.3 or $\beta$ is in the range of 1 to 1.7. The results show that our method is insensitive to fusion parameters.

%--------------------------------------------------------------------------

\subsection{Ablation Study}
\vspace{-0.15cm}
The ablation study on the instance-level state information of pedestrians and the decoupling of spatial-temporal patterns is presented in this part.

Four protocols are taken into consideration.
The first one is the proposed method itself.
In the second protocol, $p_{spa}$ and $p_{tem}$ in Eq.~\ref{equ:fusion1} are replaced by $p_{i,j}$ (Eq.~\ref{equ:spatial2}) and $f_{i,j}(\delta)$ (Eq.~\ref{equ:parzen1}). The instance-level state information is excluded in this protocol.

In the third protocol, the spatial pattern and temporal pattern are coupled together. The normalized factor in Eq.~\ref{equ:parzen3} is replaced by $\hat{Z}$:
\begin{equation}
\hat{f}^s_{i,j}(\delta) = \frac{1}{\hat{Z}^s} \sum_n H^s_{i,j}(\delta) K(n-\delta) 
\label{equ:normal_rep1}
\end{equation}
\begin{equation}
\hat{Z}^s = \max_{i,j}Z^s_{i,j} 
\label{equ:normal_rep2}
\end{equation}
which means all time interval distributions share an identical denominator. 
The numerical relations of the area under curves indicate the transmission probabilities between cameras. 
And $p_{spa}$ and $p_{spa}$ are replaced by $p_{st}$ Eq.~\ref{equ:fusion3}:
\begin{equation}
\mathcal{P} = \frac{1}{1+e^{ \beta p_{st}}},
\label{equ:normal_rep3}
\end{equation}
\begin{equation}
p_{st} = \hat{f}^s_{i,j}(\delta)
\label{equ:normal_rep4}
\end{equation}

In the fourth protocol, the instance-level state information is excluded based on the third protocol:
\begin{equation}
\hat{f}_{i,j}(\delta) = \frac{1}{\hat{Z}} \sum_n H_{i,j}(\delta) K(n-\delta) 
\label{equ:abl41}
\end{equation}
\begin{equation}
\hat{Z} = \max_{i,j}Z_{i,j} 
\label{equ:abl42}
\end{equation}

The results of these four protocols are shown in Tab.~\ref{tab:abl}. The results show that the instance-level state information of pedestrians and the decoupling of spatial and temporal are both useful to improve the performance of person re-identification.

\begin{table}[t]
\begin{center}
\setlength{\tabcolsep}{1mm}
{
\begin{tabular}{c | c | c | c | c}
\hline
\bf{Protocol} & \bf{Instance Info.} & \bf{ST Decouple} & \bf{mAP} & \bf{Rank-1}  \\
\hline

1 &\checkmark & \checkmark & \bf{89.1\%} & \bf{95.7\%} \\
2 & $\times$ & \checkmark & 87.1\% & 94.3\% \\
3 & \checkmark & $\times$ & 86.9\% & 95.0\% \\
4 & $\times$ & $\times$ & 83.4\% & 93.8\% \\

\hline
\end{tabular}}
\end{center}
\caption{Ablation Study results on DukeMTMC-reID~\cite{Ergys2016Performance}.}
\label{tab:abl}
\vspace{-0.3cm}
\end{table}

Comparing the second and the fourth protocol, the mAP and Rank-1 accuracy are improved by 3.7\% and 0.5\%.
Comparing the third and the fourth protocol, the mAP and Rank-1 accuracy are improved by 3.5\% and 1.2\%.
The instance-level state information of pedestrians is more helpful to Rank-1 accuracy, and the decoupling of spatial and temporal patterns is more contributive to mAP.
These results indicate that the decoupling of spatial and temporal patterns is more helpful to improve mAP by recalling more hard positive samples. 
The instance-level state information of pedestrians is more helpful to improve precision by narrow the number of gallery images.
The combination of these two strategies achieves the best performance.

\begin{figure}[t]
\begin{center}
   \includegraphics[width=\linewidth]{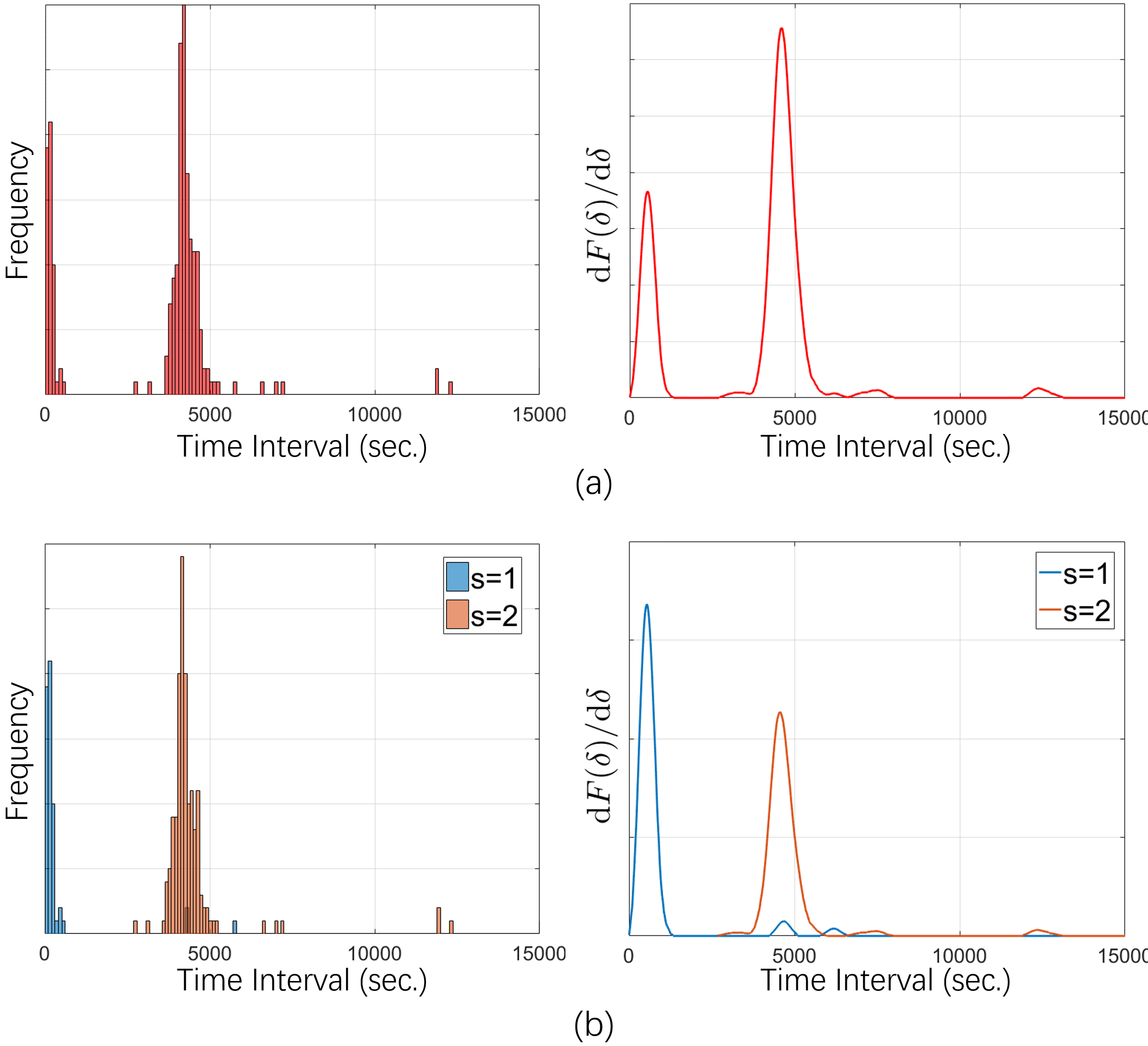}
\end{center}
\vspace{-0.3cm}
   \caption {The time interval distributions between \textit{camera 1} and \textit{camera 2} of DukeMTMC-reID. \textbf{(a)}: Time interval distribution without state information. \textbf{(b)}: Time interval distributions with instance-level state information. The distribution in \textbf{(a)} is split into two distributions in \textbf{(b)}, which means more irrelevant gallery images can be filtered out according to the instance-level state information.}
\label{fig:time_dis2}
\vspace{-0.3cm}
\end{figure}

To demonstrate the effect of introducing instance-level state information, the time interval distributions between \textit{camera 1} and \textit{camera 2} of DukeMTMC-reID are shown in Fig.~\ref{fig:time_dis2}. The instance-level states are not taken into consideration in Fig.~\ref{fig:time_dis2} (a). On the other hand, distributions of two states are shown separately in Fig.~\ref{fig:time_dis2} (b). 
The distribution in Fig.~\ref{fig:time_dis2} (a) is split into two distributions, which means more irrelevant gallery images can be filtered out according to the instance-level state information.

\begin{figure}[t]
\begin{center}
   \includegraphics[width=\linewidth]{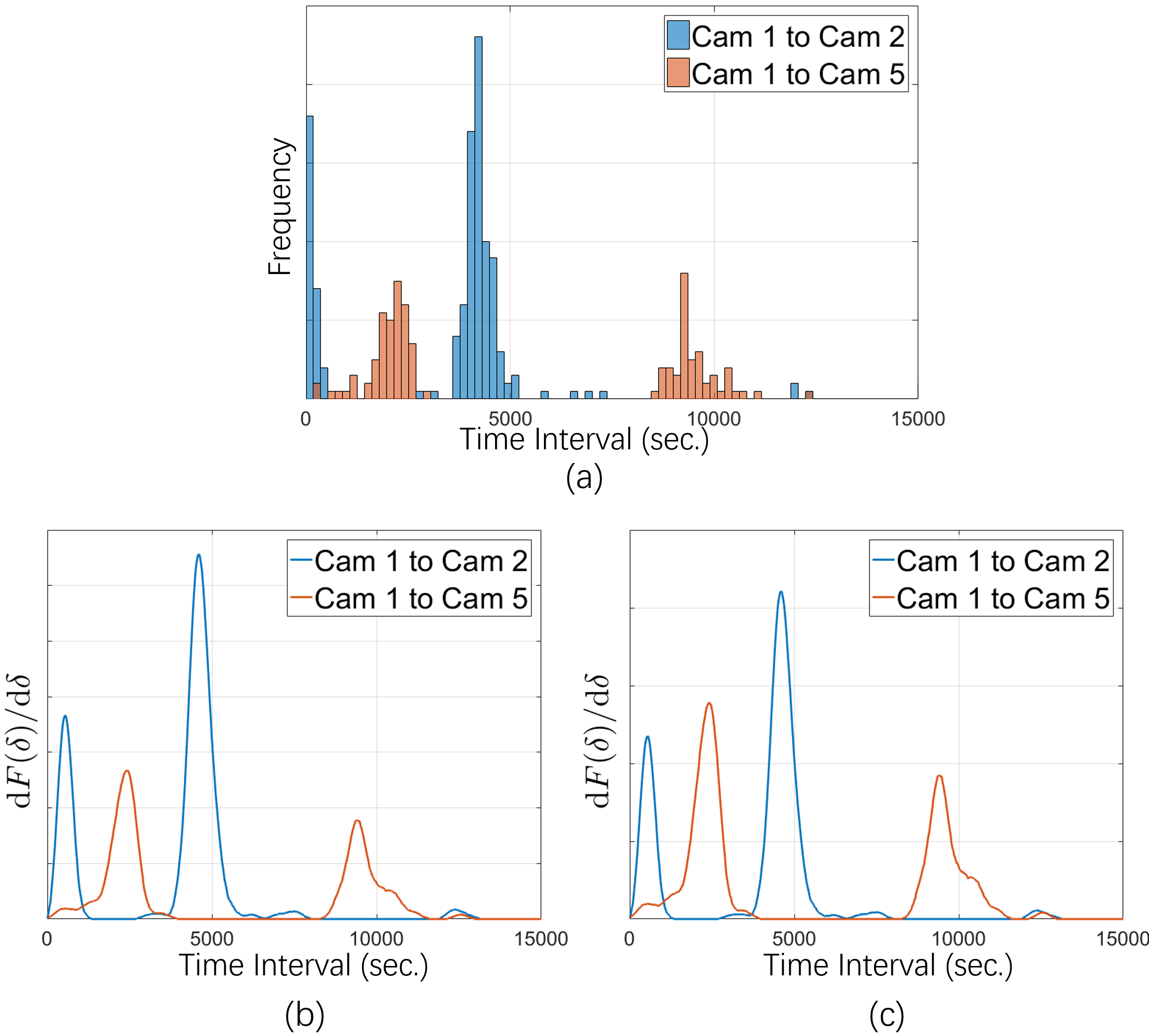}
\end{center}
\vspace{-0.3cm}
   \caption {The time interval distributions of \textit{camera 1} to \textit{camera 2} and \textit{camera 1} to \textit{camera 5} of DukeMTMC-reID. \textbf{(a)}: Time interval frequencies. \textbf{(b)}: Time interval distributions without spatial-temporal decouple. The spatial pattern is conveyed by the areas under distribution curves. \textbf{(c)}: Time interval distributions with spatial-temporal decouple. The areas under distribution curves are the same. The spatial pattern is decoupled from the time interval distributions as transmission probabilities. (Instance-level state information is not shown for simplicity.)}
\label{fig:time_dis1}
\vspace{-0.4cm}
\end{figure}

To show the difference between spatial-temporal coupled constraint and spatial-temporal decoupled constraint, time interval distributions of \textit{camera 1} to \textit{camera 2} and \textit{camera 1} to \textit{camera 5} are shown in Fig.~\ref{fig:time_dis1}. In the coupled case, the spatial pattern is conveyed by the areas under distribution curves as shown in Fig.~\ref{fig:time_dis1} (b). On the contrary, in the decoupled case, the areas under distribution curves are the same  as shown in Fig.~\ref{fig:time_dis1} (c), and the spatial pattern is decoupled from the time interval distributions as transmission probabilities.

%--------------------------------------------------------------------------

\subsection{Failure Analysis}
\vspace{-0.15cm}
In this part, we analyze the failure cases of the proposed method on DukeMTMC-reID.
We find that the failures can be categorized as four cases:

Firstly, there are incorrect labels in DukeMTMC-reID. 
The proportion of failure cases caused by incorrect labels is 16.2\%.
It is harmful to keep the incorrect labels in the database. \textbf{Hence, we release a cleaned data list of DukeMTMC-reID with this paper: \url{https://github.com/RenMin1991/cleaned-DukeMTMC-reID/}}.

%\begin{figure}[h]
%\begin{center}
%   \includegraphics[width=0.9\linewidth]{figures/fail1.png}
%\end{center}
%  \caption {Failure case 1: there are incorrect labels in DukeMTMC-reID. The proportion of this case in all failures is 16.2\%.}
%\label{fig:fail1}
%\end{figure}

%\begin{figure}[h]
%\begin{center}
%   \includegraphics[width=\linewidth]{figures/fail2.png}
%\end{center}
%   \caption {Failure case 2: the pedestrians are occluded seriously. The proportion of this case in all failures is 56.9\%.}
%\label{fig:fail2}
%\end{figure}

Secondly, the feature extractor is fooled because of serious occlusions. For example, the upper part of two individuals is quite similar while the lower part is occluded.
The proportion of this case in all failures is 56.9\%.

In the third case, the visual feature is not discriminative enough to distinguish the hard negative samples.
The proportion of this case in all failures is 23.5\%.

%\begin{figure}[h]
%\begin{center}
%   \includegraphics[width=0.9\linewidth]{figures/fail34.png}
%\end{center}
%   \caption {\textbf{(a)} Failure case 3: the visual feature is not discriminative enough. The proportion of this case in all failures is 23.5\%. \textbf{(b)} Failure case 4: the proposed method improperly pushes up the final joint metric. The proportion of this case in all failures is 3.4\%.}
%\label{fig:fail34}
%\end{figure}

In the last case, the proposed method outputs high probabilities due to spatial-temporal patterns. However, it improperly pushes up the final joint metric.
The proportion of this case in all failures is 3.4\%.

The failure analysis shows that serious occlusion is the main cause of mismatching (56.9\%).
The second important reason is that the feature extracted by the recognition model is not discriminative enough for some similar images (23.5\%).
Incorrect labels also degrade the performance (16.2\%).
The proportion of failure samples caused by the improper spatial-temporal probability in all failures is quite small (3.4\%).

%%%%%%%%%%%%%%%%%%%%%%%%%%%%%%%%%%%
%%%%%%%%%%%%%%%%%%%%%%%%%%%%%%%%%%%

\section{Conclusion}
\vspace{-0.1cm}
In this paper, we propose a method to exploit spatial-temporal patterns for person re-identification.
% where the instance-level state information is taken into consideration to provide personalized predictions.
%
Different from the existing spatial-temporal person re-identification methods, the proposed method adopts the walking direction of each pedestrian, as key instance-level state information, to provide personalized predictions.
%the proposed method is established on the instance-level state information, which can provide personalized predictions for each pedestrian. 
%
In addition, the spatial-temporal patterns are decoupled into transmission probabilities and time interval distributions between cameras. The spatial-temporal patterns become mutually beneficial rather than in conflict with each other as current methods.
A novel joint metric is proposed to fuse the instance-level spatial constraint, temporal constraint, and visual feature similarity.
The superiority of our method is demonstrated by extensive contrast experiments.
And adequate experimental analyses provide more insights into our method.

{\small
\bibliographystyle{ieee_fullname}
\bibliography{egpaper_for_review}

\begin{thebibliography}{10}\itemsep=-1pt

\bibitem{Hermans2017In}
Hermans Alexander, Beyer Lucas, and Leibe Bastian.
\newblock In defense of the triplet loss for person re-identification.
\newblock {\em ArXiv 1703.07737}, 2017.

\bibitem{Apurva2011Multiple}
Bedagkar-Gala Apurva and Shah~Shishir K.
\newblock Multiple person re-identification using part based spatio-temporal
  color appearance model.
\newblock In {\em IEEE International Conference on Computer Vision Workshops
  (ICCV Workshops)}, 2011.

\bibitem{Su2017Learning}
Su C., Li J., Zhang S., Xing J., Gao W., and Q. Tian.
\newblock Pose-driven deep convolutional model for person reidentification.
\newblock {\em IEEE/CVF International Conference on Computer Vision (ICCV)},
  2017.

\bibitem{Song2018Mask}
Song Chunfeng, Huang Yan, Ouyang Wanli, and Wang Liang.
\newblock Mask-guided contrastive attention model for person re-identification.
\newblock In {\em IEEE/CVF Conference on Computer Vision and Pattern
  Recognition (CVPR)}, 2018.

\bibitem{2015An}
Ahmed Ejaz, Jones Michael, and Tim~K. Marks.
\newblock An improved deep learning architecture for person re-identification.
\newblock In {\em IEEE Conference on Computer Vision \& Pattern Recognition
  (CVPR)}, pages 3908--3916, 2015.

\bibitem{Ergys2016Performance}
Ristani Ergys, Solera Francesco, Zou~Roger S, Cucchiara Rita, and Tomasi Carlo.
\newblock Performance measures and a data set for multi-target, multi-camera
  tracking.
\newblock In {\em European Conference on Computer Vision (ECCV)}, 2016.

\bibitem{guangcong2019aaai}
Wang Guangcong, Lai Jianhuang, Huang Peigen, and Xie Xiaohua.
\newblock Spatial-temporal person re-identification.
\newblock {\em Proceedings of the AAAI Conference on Artificial Intelligence},
  pages 8933--8940, 2019.

\bibitem{Wang2017P2SNet}
Wang Guangcong, Lai Jianhuang, and Xie Xiaohua.
\newblock P2snet: Can an image match a video for person re-identification in an
  end-to-end way?
\newblock {\em IEEE Transactions on Circuits and Systems for Video Technology},
  2017.

\bibitem{Wang2016DARI}
Wang Guangrun, Lin Liang, Ding Shengyong, Li Ya, and Wang Qing.
\newblock Dari: Distance metric and representation integration for person
  verification.
\newblock {\em Proceedings of the AAAI Conference on Artificial Intelligence},
  2016.

\bibitem{2018Learning}
Wang Guanshuo, Yuan Yufeng, Chen Xiong, Li Jiwei, and Zhou Xi.
\newblock Learning discriminative features with multiple granularities for
  person re-identification.
\newblock {\em 2018 ACM Multimedia Conference}, 2018.

\bibitem{Luo2019Bag}
Luo Hao, Gu Youzhi, Liao Xingyu, Lai Shenqi, and Jiang Wei.
\newblock Bag of tricks and a strong baseline for deep person
  re-identification.
\newblock In {\em The IEEE Conference on Computer Vision and Pattern
  Recognition Workshops (CVPR Workshops)}, June 2019.

\bibitem{He2018CVPR}
Lingxiao He, Jian Liang, Haiqing Li, and Zhenan Sun.
\newblock Deep spatial feature reconstruction for partial person
  re-identification: Alignment-free approach.
\newblock In {\em Proceedings of the IEEE Conference on Computer Vision and
  Pattern Recognition (CVPR)}, June 2018.

\bibitem{he2020fastreid}
Lingxiao He, Xingyu Liao, Wu Liu, Xinchen Liu, Peng Cheng, and Tao Mei.
\newblock Fastreid: A pytorch toolbox for general instance re-identification.
\newblock {\em arXiv preprint arXiv:2006.02631}, 2020.

\bibitem{2020Foreground}
Lingxiao He, Wang Yinggang, Liu Wu, Zhao He, Sun Zhenan, and Feng Jiashi.
\newblock Foreground-aware pyramid reconstruction for alignment-free occluded
  person re-identification.
\newblock In {\em IEEE/CVF International Conference on Computer Vision (ICCV)},
  2019.

\bibitem{2018Unsupervised}
Lv Jianming, Chen Weihang, Li Qing, and Yang Can.
\newblock Unsupervised cross-dataset person re-identification by transfer
  learning of spatial-temporal patterns.
\newblock {\em IEEE Conference on Computer Vision \& Pattern Recognition
  (CVPR)}, 2018.

\bibitem{2019Pose}
Miao Jiaxu, Wu Yu, Liu Ping, Ding Yuhang, and Yang Yi.
\newblock Pose-guided feature alignment for occluded person re-identification.
\newblock In {\em 2019 IEEE/CVF International Conference on Computer Vision
  (ICCV)}, 2019.

\bibitem{2018Pose}
Liu Jinxian, Ni Bingbing, Yan Yichao, Zhou Peng, and Hu Jianguo.
\newblock Pose transferrable person re-identification.
\newblock In {\em IEEE/CVF Conference on Computer Vision and Pattern
  Recognition (CVPR)}, 2018.

\bibitem{2017Joint}
Cho~Yeong Jun, Kim~Su A, Park~Jae Han, Lee Kyuewang, and Yoon~Kuk Jin.
\newblock Joint person re-identification and camera network topology inference
  in multiple cameras.
\newblock {\em Computer Vision and Image Understanding}, 2017.

\bibitem{2018MaskReID}
Qi Lei, Huo Jing, Wang Lei, Shi Yinghuan, and Gao Yang.
\newblock Maskreid: A mask based deep ranking neural network for person
  re-identification.
\newblock {\em ArXiv 1804.03864}, 2018.

\bibitem{Zheng2015Scalable}
Zheng Liang, Shen Liyue, Tian Lu, Wang Shengjin, and Tian Qi.
\newblock Scalable person re-identification: A benchmark.
\newblock In {\em IEEE International Conference on Computer Vision (ICCV)},
  2015.

\bibitem{Kalayeh2018Human}
Kalayeh~Mahdi M., Basaran Emrah, Gokmen Muhittin, Kamasak~Mustafa E., and Shah
  Mubarak.
\newblock Human semantic parsing for person re-identification.
\newblock {\em IEEE/CVF Conference on Computer Vision and Pattern Recognition
  (CVPR)}, 2018.

\bibitem{Ma2018Covariance}
Bingpeng Ma, Yu Su, and Frederic Jurie.
\newblock Covariance descriptor based on bio-inspired features for person
  reidentification and face verification.
\newblock {\em Image and Vision Computing}, 2014.

\bibitem{2015Learning}
Sakrapee Paisitkriangkrai, Chunhua Shen, and Anton Van~Den Hengel.
\newblock Learning to rank in person re-identification with metric ensembles.
\newblock {\em Proceedings of the IEEE/CVF Conference on Computer Vision and
  Pattern Recognition (CVPR)}, 2015.

\bibitem{2017A}
Sarfraz~M. Saquib, Schumann Arne, Eberle Andreas, and Stiefelhagen Rainer.
\newblock A pose-sensitive embedding for person re-identification with expanded
  cross neighborhood re-ranking.
\newblock {\em IEEE/CVF Conference on Computer Vision and Pattern Recognition
  (CVPR)}, 2018.

\bibitem{Gong2014Person}
Gong Shaogang, Cristani Marco, Yan Shuicheng, and Loy~Chen Change.
\newblock Person re-identification.
\newblock {\em Advances in Computer Vision \& Pattern Recognition},
  42(7):301--313, 2014.

\bibitem{Shen2018PersonRW}
Y. Shen, Hongsheng Li, Shuai Yi, Dapeng Chen, and X. Wang.
\newblock Person re-identification with deep similarity-guided graph neural
  network.
\newblock {\em ArXiv}, abs/1807.09975, 2018.

\bibitem{Ding2015Deep}
Ding Shengyong, Lin Liang, Wang Guangrun, and Chao Hongyang.
\newblock Deep feature learning with relative distance comparison for person
  re-identification.
\newblock {\em Pattern Recognition}, 48(10):2993--3003, 2015.

\bibitem{2016Person}
Dapeng Tao, Yanan Guo, Mingli Song, Yaotang Li, Zhengtao Yu, and Yuan~Yan Tang.
\newblock Person re-identification by dual-regularized kiss metric learning.
\newblock {\em IEEE Transactions on Image Processing}, 25(6):2726--2738, 2016.

\bibitem{Wang2020High}
Guan’an Wang, Shuo Yang, Huanyu Liu, Zhicheng Wang, Yang Yang, Shuliang Wang,
  Gang Yu, Erjin Zhou, and Jian Sun.
\newblock High-order information matters: Learning relation and topology for
  occluded person re-identification.
\newblock In {\em Proceedings of the IEEE/CVF Conference on Computer Vision and
  Pattern Recognition (CVPR)}, June 2020.

\bibitem{2016Camera}
Huang Wenxin, Hu Ruimin, Liang Chao, Yu Yi, and Zhang Chunjie.
\newblock Camera network based person re-identification by leveraging
  spatial-temporal constraint and multiple cameras relations.
\newblock In {\em International Conference on Multimedia Modeling}, 2016.

\bibitem{Lin2017Deep}
Lin Wu, Chunhua Shen, Anton Van, and Den Hengel.
\newblock Deep linear discriminant analysis on fisher networks: A hybrid
  architecture for person re-identification.
\newblock {\em Pattern Recognition}, 2017.

\bibitem{Chen2020Salience}
Chen Xuesong, Fu Canmiao, Zhao Yong, Zheng Feng, Song Jingkuan, Ji Rongrong,
  and Yang Yi.
\newblock Salience-guided cascaded suppression network for person
  re-identification.
\newblock In {\em Proceedings of the IEEE/CVF Conference on Computer Vision and
  Pattern Recognition (CVPR)}, June 2020.

\bibitem{Yang2014Salient}
Yang Yang, Yang Jimei, Yan Junjie, Liao Shengcai, and Li~Stan Z.
\newblock Salient color names for person re-identification.
\newblock {\em Europeon Conference on Computer Vision (ECCV)}, 2014.

\bibitem{2017Beyond}
Sun Yifan, Zheng Liang, Yang Yi, Tian Qi, and Wang Shengjin.
\newblock Beyond part models: Person retrieval with refined part pooling (and a
  strong convolutional baseline).
\newblock {\em Europeon Conference on Computer Vision (ECCV)}, 2018.

\bibitem{Perceive2020Perceive}
Sun Yifan, Xu Qin, Li Yali, Zhang Chi, Li Yikang, Wang Shengjin, and Sun Jian.
\newblock Perceive where to focus: Learning visibility-aware part-level
  features for partial person re-identification.
\newblock {\em Proceedings of the IEEE/CVF Conference on Computer Vision and
  Pattern Recognition (CVPR)}, 2019.

\bibitem{Zhang2020Relation}
Zhizheng Zhang, Cuiling Lan, Wenjun Zeng, Xin Jin, and Zhibo Chen.
\newblock Relation-aware global attention for person re-identification.
\newblock In {\em Proceedings of the IEEE/CVF Conference on Computer Vision and
  Pattern Recognition (CVPR)}, June 2020.

\bibitem{2016Deep}
Chen~Shi Zhe, Guo~Chun Chao, and Lai Jianhuang.
\newblock Deep ranking for person re-identification via joint representation
  learning.
\newblock {\em IEEE Transactions on Image Processing}, pages 1--1, 2016.

\bibitem{Zheng2018Person}
Liang Zheng, Yi Yang, and Alexander~G Hauptmann.
\newblock Person re-identification: Past, present and future.
\newblock {\em ArXiv}, 1610.02984, 2016.

\end{thebibliography}
}

\end{document}